\documentclass[sigconf,nonacm]{acmart}
\renewcommand\footnotetextcopyrightpermission[1]{}
\usepackage{booktabs}
\usepackage{amsmath}
\usepackage{tabularx}

\AtBeginDocument{}

\newtheorem{lemma}{Lemma}
\newtheorem{proposition}{Proposition}
\newtheorem{definition}{Definition}

\newcommand{\Eb}{\mathbb{E}}
\newcommand{\Pb}{\mathbb{P}}
\newcommand{\sig}{\sigma}
\newcommand{\coverageentry}[3]{#1(#2)\,[#3]}

\begin{document}

\title[Which Loans, Not How Many]{Which Loans, Not How Many: Exactly
Level-Invariant, Anytime-Valid Monitoring of Deployed Risk Models}

\author{Xin Xu}
\email{xuxin@cmu.edu}
\affiliation{\institution{Carnegie Mellon University}\city{Pittsburgh}\state{PA}\country{USA}}

\begin{abstract}
A refinancing wave can change the number of mortgage prepayments
without changing the relationship between scores and prepayment.
In a logistic model, conditioning on the monthly count removes the
intercept from the likelihood of which loans prepay. Our fixed-start
slope product tests a lower bound on the calibration slope. At threshold
20 its probability of ever crossing under the null is at most 5\%.
The restarted level sum tracks departures from intercept tolerance.
A confidence sequence estimates a constant slope, and the
restart-intersection rule detects changes in that slope. A continuous
mixture avoids the off-grid width floor of a discrete prior.
On Freddie Mac loans, fixed-slope resampling reproduces 78\% of
maintained-model monthly AUC variance in development and 70\% in
the holdout (95\% intervals 57--100\% and 48--100\%). From fresh
holdout starts, the restarted level sum crosses on both paths,
and the fixed-start slope product stays below threshold on both.
The restart-intersection rule signals a downward change for the frozen
model in April 2018 and an upward change for the maintained model
in a supplementary run in August 2019.
\end{abstract}



\maketitle

\section{Introduction}\label{sec:intro}

When mortgage rates fall, more borrowers refinance, and a portfolio can
prepay faster than its model predicts. A monitoring team then has to decide
whether updating the predicted level is enough or whether the association
between the model's scores and prepayment has changed. The distinction
matters for the model reviews required in banking
\cite{frb2026,ecb2019,siddiqi2017}.

Write $\theta_{it}=\operatorname{logit}\hat p_{it}$ for the model's scored
log-odds for loan $i$ in month $t$. The logistic recalibration model of
Cox \cite{cox1958,miller1991} gives the prepayment probability
\begin{equation}\label{eq:model}
\Pb(Y_{it}=1)=\sig(\alpha_t+\beta_t\theta_{it}),
\end{equation}
where $\alpha_t$ is the calibration intercept and $\beta_t$ is the
calibration slope \cite{vancalster2019}. The slope measures association per
unit of stated log-odds. Attainable AUC also depends on the intercept and
the score distribution (Appendix~\ref{app:auc}), so AUC alone cannot
isolate changes in that association. Multiplying the logits by a positive
constant $c$ divides the slope by $c$ and preserves the ranking
\cite{stevens2020}.

Given the past, the current book and the month's intercept, we model
loan outcomes as independent. Conditional on the observed count, the
likelihood of which loans prepay depends only on the slope. We use
three monitors. The \emph{fixed-start slope product} multiplies monthly
conditional likelihood ratios to test a lower bound on the slope.
At threshold 20 its probability of ever crossing under the null is
at most $1/20$. The \emph{restarted level sum} tracks departures from
intercept tolerance on the model's stated scale. The
\emph{restart-intersection rule} intersects slope confidence sequences
started at successive months and signals when they have no slope in
common. The newer binding set gives the direction of change.

The application follows maintained and frozen prepayment models on a
Freddie Mac panel spanning 1999--2025. Resampling the observed books
and counts at a fixed slope reproduces 78\% of the maintained model's
monthly AUC variance in development and 70\% in the holdout. From fresh
holdout starts, both paths cross the level threshold and both
fixed-start slope products stay below threshold. The restart-intersection
rule signals a downward change on the frozen path and an upward change
in a supplementary maintained run.
In simulations a continuous mixture avoids the off-grid width floor of
a discrete slope prior. On the Freddie Mac panel, the grid and
continuous-set implementations agree on the five development and frozen
holdout change dates.

\section{Related Work}\label{sec:related}

\textbf{Anytime-valid monitoring.}
E-values and confidence sequences \cite{ville1939,howard2021,ramdas2023}
have been used to monitor predictive risk, calibration, fairness and
forecast performance \cite{podkopaev2022,chugg2023,choe2024},
including settings with interventions \cite{feng2024}.
\citet{arnold2023} use the probability integral transform
(PIT) to test probabilistic forecast calibration sequentially.
Farran's PITMonitor \cite{farran2026} uses a mixture e-process on PITs
and estimates the changepoint. \citet{wangEbacktesting2026}
use e-processes for sequential rejection of risk forecasts, and \citet{henzi2023}
test binary probability calibration in a credit
application. Our monitor targets the calibration slope on a specified
log-odds scale. Its intercept invariance comes from conditioning on the
monthly count before combining the likelihoods
(Table~\ref{tab:related-comparison}).
Likelihood-ratio confidence sequences for generalized linear models
\cite{lee2024} cover a constant joint parameter.

\textbf{Conditional e-values and change detection.}
The per-month factor is a conditional (COND) e-variable
\cite{hao2024}. Hao and Gr\"unwald study fixed-sample e-power.
Here the conditioning takes place separately within each month, so the
conditional mean bound composes across months. Turner and Gr\"unwald
\cite{turner2023} construct confidence sequences for $2\times2$ effect sizes
using reverse information projection. Our conditional likelihood uses
continuous scores and includes the odds-ratio model for binary scores as a
special case. We use restart constructions from sequential change detection
\cite{shin2024,shekhar2024}.

\textbf{Prediction-model validation.}
Case mix can change discrimination measures
\cite{morise1996,vergouwe2010,vanamsterdam2024}. Model-based concordance
$C^M$ \cite{vanklaveren2016,vanklaveren2019} and model-based ROC curves
\cite{sadatsafavi2022} provide references for such shifts. Wang
\cite{wang2026} uses counterfactual comparisons to assess how much of a
decline in the Kolmogorov--Smirnov statistic comes from sampling variation
or portfolio shifts. Monitoring
practice also uses cumulative accuracy profiles, AUC, Gini, stability
indices and calibration checks \cite{bcbs2005,siddiqi2017,casellina2026}.
We compare with the ECB statistic, model-based concordance and monthly
joint logistic recalibration. We also fit the slope with the intercept at
its commissioned value to measure how a level shift affects that fit.

\section{The conditional reduction}\label{sec:reduction}

Fix a month $t$ with monitored book $\{\theta_i\}_{i=1}^{n}$; scores are
out-of-sample (the model producing $\hat p$ is trained only on data before the
month's training cutoff). Let $Y_i\sim\mathrm{Bern}(\sig(\alpha+\beta\theta_i))$
independently within the month, $S=\{i:Y_i=1\}$, $M=|S|$, and
$T=\sum_{i\in S}\theta_i$.

\begin{lemma}[classical \cite{cox1958,cox1970,chen1994,hirji1987}]
\label{lem:factor}
$\Pb_{\alpha,\beta}(S)=\Pb_{\alpha,\beta}(M)\cdot\Pb_{\beta}(S\mid M)$ with
\begin{equation}\label{eq:arrangement}
\Pb_{\beta}(S\mid M{=}m)=\frac{e^{\beta T}}{Z_m(\beta)},\quad
Z_m(\beta)=\!\!\sum_{|A|=m}\exp\Big(\beta\!\sum_{i\in A}\theta_i\Big),
\end{equation}
the conditional-Bernoulli law \cite{chen1994,chen1997}: a one-parameter
exponential family in $\beta$ with sufficient statistic $T$, free of
$\alpha$. The count factor is an exponential family in $\alpha$ with
sufficient statistic $M$ and log-normalizer
$\sum_i\log(1+e^{\alpha+\beta\theta_i})$.
\end{lemma}

Conditioning on $M$ removes $\alpha$ because its likelihood factor is
common to all event sets of size $M$. For each slope value, a log-space
elementary-symmetric recursion computes $\log Z_m(\beta)$ for all counts
through $m_{\max}$ in $O(nm_{\max})$ time.

\section{Anytime-valid inference for the slope}\label{sec:av}

Fix an error level $0<\alpha<1$. Let $\mathcal F_t$ contain the outcomes
through month $t$ and the information available before outcomes in
months $1,\ldots,t+1$, including commissioning data and the next book.
The month-$t$ book and $\beta_t$ are $\mathcal F_{t-1}$-measurable.
Conditional on $\mathcal F_{t-1}$ and the current month's intercept
$\alpha_t$, the book's outcomes $Y_{it}$ are independent Bernoulli
variables with probabilities $\sig(\alpha_t+\beta_t\theta_{it})$.
The intercept may be unobserved.
The conditional likelihood ratio for month $t$ is
\[
E_t(\beta_1,\beta_0)=e^{(\beta_1-\beta_0)T_t}\,
Z_{M_t}(\beta_0)/Z_{M_t}(\beta_1).
\]

\begin{proposition}[e-process under the composite null]\label{prop:eprocess}
Fix $\beta_0$ at time zero. Let $E_t$ be the factor at a predictable
$\beta_1<\beta_0$ or a probability mixture of such factors with weights
chosen before month $t$.
Under $H_0:\beta_t\ge\beta_0$ for all $t$,
$\Eb[E_t\mid\mathcal F_{t-1},M_t]\le1$. Thus $R_T=\prod_{t\le T}E_t$
is a nonnegative supermartingale with $R_0=1$ and
$\Pb(\sup_T R_T\ge1/\alpha)\le\alpha$.
\end{proposition}

The proof is in Appendix~\ref{app:proofs}.

\begin{proposition}[confidence sequence]\label{prop:cs}
For a fixed probability measure $\Pi$ on the slope domain, define
$A_T(\beta)=\sum_{t\le T:\,M_t\ge5}\{\beta T_t-\log Z_{M_t}(\beta)\}$ and
\[
\log\mathcal M_T(\beta_0)=\log\int e^{A_T(b)}\,\Pi(db)-A_T(\beta_0).
\]
If $\beta_t\equiv\beta^\star$ in the slope domain, the process
$\mathcal M_T(\beta^\star)$ is a nonnegative martingale with mean one, and the set
$\mathrm{CS}_T=\{\beta_0:\mathcal M_T(\beta_0)<1/\alpha\}$ satisfies
$\Pb(\exists T:\beta^\star\notin\mathrm{CS}_T)\le\alpha$.
\end{proposition}

The proof is in Appendix~\ref{app:proofs}.

\textbf{Restart-intersection rule.}
At month $t$ we intersect the exact sets $\mathrm{CS}_{r,t}$ from
Proposition~\ref{prop:cs} started at each $r\le t$ using months $r$ through
$t$. The alarm time $\tau$ is the first month with an empty intersection.
With a constant slope in the domain under the monthly conditional
model, the exact-set rule has a mean time to false alarm of at least
$1/\alpha$, or 20 months at $\alpha=0.05$ \cite{shekhar2024}. The
intersection is empty when the largest lower endpoint exceeds the
smallest upper endpoint, up to equality at the acceptance boundary.
When the binding sets have distinct starts, we record the alarm as
upward if the newer set lies above the older set and downward if it
lies below. Appendix~\ref{app:proofs} derives the run-length bound.

We recommend a continuous uniform prior on $[0.20,1.60]$. Between
likelihood evaluations, concavity of $A_T$ bounds it below by its chord.
Integrating the exponentiated chords gives a lower bound on the mixture
numerator and hence an outer confidence set. The integration mesh can
be refined at each look to control the numerical gap. The mortgage-panel
sequences and intersections use an equal-weight 29-node prior on the
same range (Appendix~\ref{app:numerics}).

\textbf{Restarted level sum.}
The count factor gives a likelihood for $\alpha$ after fixing $\beta$, with
log-normalizer $\log C(\alpha,\beta)=\sum_i\log(1+e^{\alpha+\beta\theta_i})$.
The level sequence is valid under the working model with a pinned slope and
an intercept constant over its window. For the level axis we use $\beta=1$,
the model's stated log-odds scale, so $\alpha=0$ corresponds to the stated
probabilities. To monitor the tolerance $|\alpha_t|\le 0.10$, we
average monthly count likelihood ratios for alternatives beyond
either boundary against the corresponding boundary, then sum the
products of these factors over all start months. The
restarted level sum signals at 20.

\section{A pinned-slope excess-variance diagnostic}\label{sec:rho}

\begin{definition}\label{def:rho}
For a monthly monitoring statistic $S_t$, any function of the month's
book and event set with positive observed variance, the
\emph{pinned-slope excess-variance share} is
\[
\rho(S)=1-\operatorname{Var}_{H_0}[S]\,/\operatorname{Var}_{\mathrm{obs}}[S],
\]
where $\operatorname{Var}_{H_0}$ is the across-month variance when each
month's event set is replaced by a draw from \eqref{eq:arrangement} at a
pinned $\beta$, holding the real book and the real count.
\end{definition}

The resampling uses exact conditional draws, independently across months.
Count-only and book-only statistics are unchanged draw by draw and have
$\rho=0$ when their observed variance is positive. The raw contrast can
be negative, and the displayed shares are clipped at zero.
The first-difference version $\rho_\Delta$ measures
excess variation between successive months. Excess can reflect slope
variation, dependence or other departures from the benchmark. Intervals use
a moving-block bootstrap over the observed months with block length 12.

\section{Experimental Design}\label{sec:data}

\textbf{Panel.}
The Freddie Mac Single-Family Loan-Level Dataset \cite{freddie}
contributes 26 vintages (1999--2024) of 50{,}000 loans each, for 1.3M loans
and 69.3M loan-months through 2025. The panel through December 2015 contains
35.6M loan-months and 534{,}724 prepayments (1.50\% per month). Macro
covariates are joined as of the prediction date, with mortgage rates and
unemployment lagged one month and FHFA house prices lagged two months
with state-level interpolation.

\textbf{Models and monitored books.}
We use LightGBM \cite{ke2017}, retrained each January on a trailing ten-year
window (maintained), and a spline-logistic GLM. Frozen variants keep the
fitted scoring function from the named year. All predictions are
out of sample. Each monthly monitored book has 20{,}000 loans. The slope
analysis requires at least five prepayments, and probabilities are clipped
to $[10^{-6},1-10^{-6}]$ before taking logits. The monitor uses the
current book's scores and outcomes.

\textbf{Windows and thresholds.}
The design used January 2001--December 2003 for commissioning and January
2004--December 2015 for development, set the absolute tolerance at
$\delta=0.10$ slope units, and held out January 2016--August 2025 until the
analysis was set. The LightGBM commissioning estimate is
$\hat\beta_{\mathrm{ref}}=0.759$ (SE 0.007), with lower endpoint
$\beta_{\mathrm{ref}}^{\mathrm{lo}}=0.746$ of the 95\% Wald interval and alarm boundary
$\beta_{\mathrm{ref}}^{\mathrm{lo}}-\delta=0.646$. The corresponding
GLM estimates are 0.828 and 0.714 for the slope and alarm boundary.
We use the common boundary 0.646 and start the fixed-start slope
product at the first scored month in each development or holdout
window, after commissioning. A crossing of 20 is the primary one-sided
alarm. The preregistered one-sided analysis used the restarted slope
sum (Table~\ref{tab:hyps}). The windows, tolerance and hypotheses were
registered before downloading loan-level data.

\textbf{Code and data availability.}
The Freddie Mac SFLLD sample is publicly available under its license.
The macroeconomic inputs come from FRED and FHFA.
Code, configuration, and outputs will be released with the published version.

\section{Results}\label{sec:results}

\begin{figure*}[t]
\centering
\includegraphics[width=0.94\textwidth]{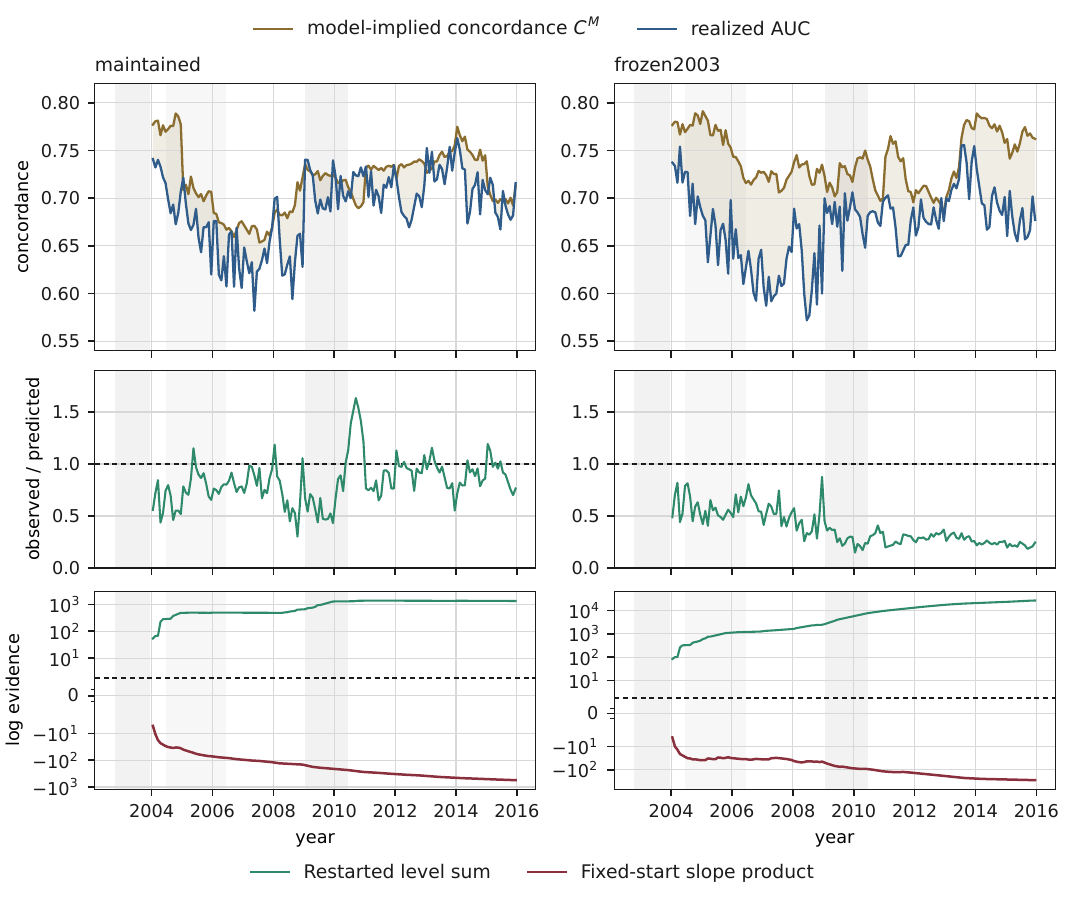}
\caption{Monthly AUC, observed-to-expected prepayments and detector
evidence, 2004--2015 (maintained model, left, and frozen2003, right).
Shading marks rising-rate and refinancing periods. The top panels compare
realized AUC with model-based concordance $C^M$. The bottom panels show
the restarted level sum and fixed-start slope product, each with
threshold 20. The fixed-start slope product stays below threshold on
both paths.}
\label{fig:headline}
\end{figure*}

\subsection{Dashboard variation under a fixed-slope benchmark}\label{sec:battery}

Figure~\ref{fig:headline} shows the monthly dashboard on both development
paths. Table~\ref{tab:battery} compares the observed variation in each
statistic with its variation under the commissioned slope in development
and in the chronological holdout.

\begin{table*}[t]
\centering
\caption{Exploratory dashboard comparison at $\beta=0.759$, with
64 redraws per month on the observed books and counts. The development
panel has 144 months per path and the holdout has 116. In development,
the Monte Carlo standard error of $\rho$ is about one percentage
point at the maintained AUC estimate.
Intervals are 95\% moving-block bootstrap intervals and displayed shares
are clipped at zero. $\Delta$ gives the share for first differences.
$\mathrm{S/N}=\sqrt{\max(\mathrm{Var}_{\mathrm{obs}}/\mathrm{Var}_{H_0}-1,0)}$.}
\label{tab:battery}
\small
\setlength{\tabcolsep}{2.4pt}
\begin{tabular}{@{}lrrrrrr@{}}
\toprule
\multicolumn{7}{@{}l}{Development}\\
& \multicolumn{3}{c}{maintained} & \multicolumn{3}{c}{frozen2003}\\
\cmidrule(lr){2-4}\cmidrule(l){5-7}
statistic & $\rho$ [95\%] & $\Delta$ & S/N & $\rho$ [95\%] & $\Delta$ & S/N\\
\midrule
realized AUC & 22 [0,43] & 0 & 0.53 & 40 [0,57] & 15 & 0.82\\
ECB $z$ (AUC vs.\ development) & 13 [0,32] & 0 & 0.39 & 0 [0,18] & 9 & 0.00\\
gap $C^M-\mathrm{AUC}$ & 60 [42,70] & 13 & 1.23 & 78 [70,83] & 37 & 1.90\\
$\hat\beta$, joint recalibration & 62 [47,70] & 26 & 1.27 & 76 [66,81] & 34 & 1.77\\
$\hat\beta$, conditional (ours) & 62 [47,70] & 26 & 1.27 & 76 [66,81] & 34 & 1.77\\
$\hat\beta$, intercept fixed (ablation) & 0 [0,31] & 2 & 0.00 & 2 [0,15] & 0 & 0.15\\
mean score of prepayers $T/M$ & 0 [0,23] & 9 & 0.00 & 17 [0,41] & 2 & 0.46\\
obs/expected; $C^M$; PSI & \multicolumn{6}{c}{$\rho\equiv0$ by construction}\\
\midrule
\multicolumn{7}{@{}l}{Holdout}\\
& \multicolumn{3}{c}{maintained} & \multicolumn{3}{c}{frozen2015}\\
\cmidrule(lr){2-4}\cmidrule(l){5-7}
statistic & $\rho$ [95\%] & $\Delta$ & S/N & $\rho$ [95\%] & $\Delta$ & S/N\\
\midrule
realized AUC & 30 [0,52] & 16 & 0.65 & 0 [0,0] & 20 & 0.00\\
ECB $z$ & 0 [0,20] & 16 & 0.00 & 0 [0,0] & 12 & 0.00\\
gap $C^M-\mathrm{AUC}$ & 43 [28,55] & 14 & 0.87 & 84 [65,87] & 29 & 2.30\\
$\hat\beta$, joint recalibration & 53 [36,62] & 0 & 1.05 & 84 [69,86] & 29 & 2.30\\
$\hat\beta$, conditional (ours) & 53 [36,62] & 0 & 1.05 & 84 [69,86] & 29 & 2.30\\
$\hat\beta$, intercept fixed (ablation) & 8 [0,41] & 2 & 0.29 & 0 [0,14] & 3 & 0.00\\
mean score of prepayers $T/M$ & 0 [0,0] & 0 & 0.00 & 0 [0,0] & 0 & 0.00\\
obs/expected; $C^M$; PSI & \multicolumn{6}{c}{$\rho\equiv0$ by construction}\\
\bottomrule
\end{tabular}
\end{table*}

\textbf{Realized AUC and the ECB statistic.}
The pinned-slope benchmark reproduces 78\% of realized AUC variance on
the maintained path (95\% interval, 57--100\%). The excess share for
first differences is 0\% after clipping (95\% interval, 0--25\%). At
alternative pins of 1 and 0.875, the benchmark reproduces 90\% and
83\% of the maintained model's AUC variance, respectively, using four
redraws per month at the 0.875 pin. On frozen2003 the excess shares are
40\% for levels (0--57\%) and 15\%
for first differences. The ECB statistic has an excess share of 13\%
(0--32\%) on maintained and 0\% (0--18\%, clipped) on frozen2003.
Its Hanley--McNeil standard error varies with the event count, so the
statistic depends on the count through its denominator as well as through
the current AUC.

In the holdout, the benchmark reproduces 70\% of maintained AUC variance
(95\% interval, 48--100\%). The corresponding shares at slope pins 1
and 0.875 are 86\% and 67\%. On frozen2015 the raw control variance is
274\% of observed AUC variance. Clipping the negative excess share
at zero gives the $0\%$ displayed in Table~\ref{tab:battery}.

\textbf{Calibration slopes and concordance.}
The conditional and joint monthly slope fits have identical excess shares
at the reported precision, 62\% on maintained and 76\% on frozen2003.
The model-based concordance gap gives 60\% and 78\%, respectively.
On the maintained path the mean score among prepayers, $T/M$, has a
clipped excess share of 0\%.

\begin{figure*}[t]
\centering
\includegraphics[width=0.94\textwidth]{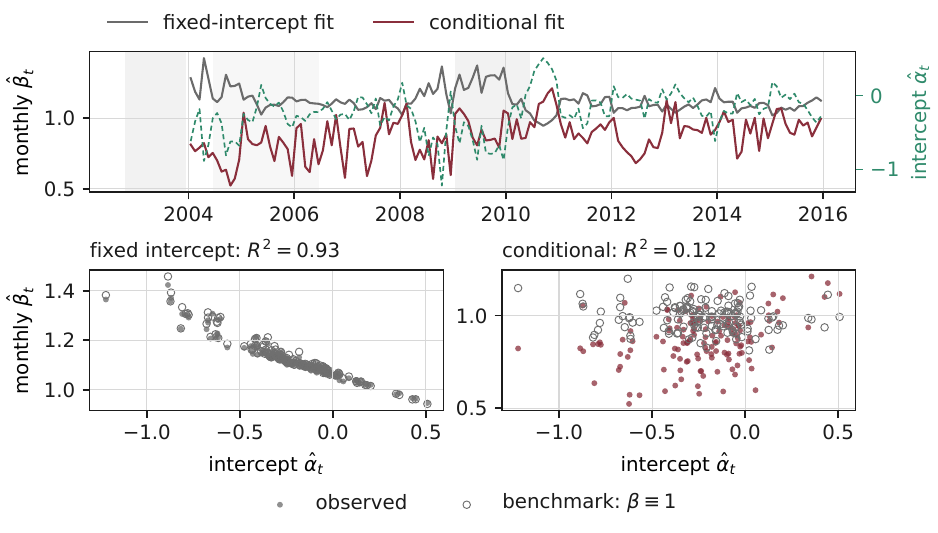}
\caption{Fixed-intercept and conditional slope fits on the maintained
path. The top panel shows the monthly fits and the estimated intercept.
The scatterplots compare observed fits (solid points) with redraws at
$\beta=1$ (open circles). The fixed-intercept fit holds $\alpha$ at its
commissioned value, 0.274.}
\label{fig:contamination}
\end{figure*}

The fixed-intercept fit correlates $-0.963$ with the estimated intercept
in the observed data and $-0.962$ in the slope-one benchmark. Its monthly
standard deviations are 0.082 and 0.086, respectively. For the conditional
fit the corresponding correlations are $+0.347$ and $-0.141$.
Figure~\ref{fig:contamination} shows how an intercept constraint can
produce apparent slope variation even at a fixed slope.
Across the 144 development months, the maintained model's conditional
slope has a Pearson correlation of $-0.364$ with the lagged 30-year
mortgage rate. The corresponding slope-one benchmark correlation is
$+0.023$.

\subsection{The development window on two axes}\label{sec:axes}

\begin{figure}[t]
\centering
\includegraphics[width=0.92\linewidth]{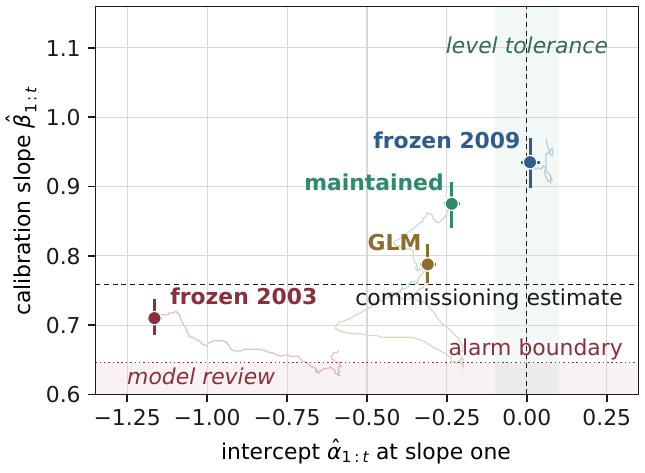}
\caption{Cumulative intercept fit at slope one (horizontal) and
conditional slope fit on the stated log-odds scale (vertical). Solid
points show terminal fits with their marginal 95\% bands.
Faint traces show the paths after burn-in. The vertical band marks the
intercept tolerance $|\hat\alpha_{1:t}|\le 0.10$. The horizontal band
lies below the common slope alarm boundary of 0.646.}
\label{fig:plane}
\end{figure}

Figure~\ref{fig:plane} plots the cumulative intercept at slope one
against the cumulative slope. When the level moves outside tolerance,
the team recalibrates the intercept. Evidence that the slope has fallen
below its boundary routes the model to review of the score--outcome
relationship.

\begin{figure*}[t]
\centering
\includegraphics[width=0.9\textwidth]{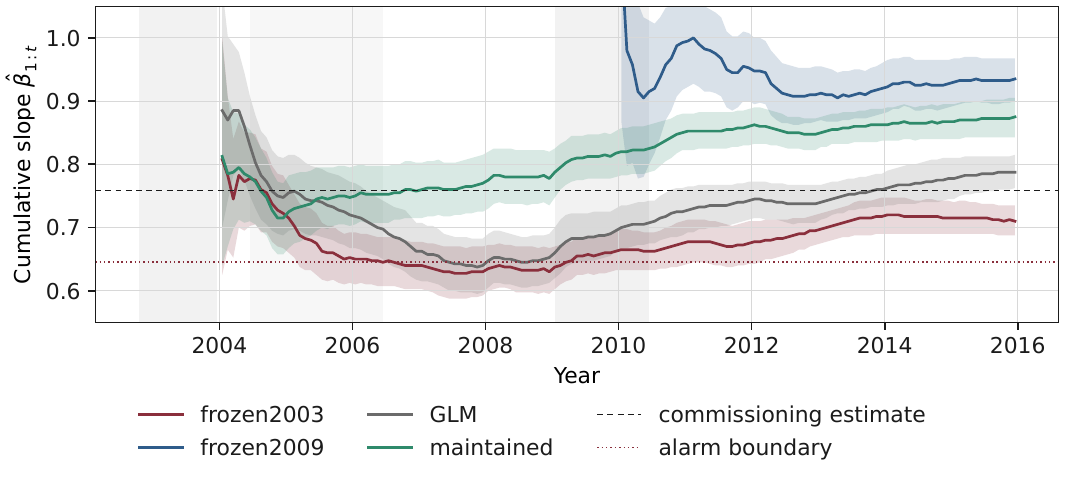}
\caption{Cumulative conditional slope fits and mixture-likelihood
bands, 2004--2015, with the commissioning estimate and alarm boundary.
Coverage in Proposition~\ref{prop:cs} assumes a constant slope.}
\label{fig:cs}
\end{figure*}

\begin{figure*}[t]
\centering
\includegraphics[width=0.94\textwidth]{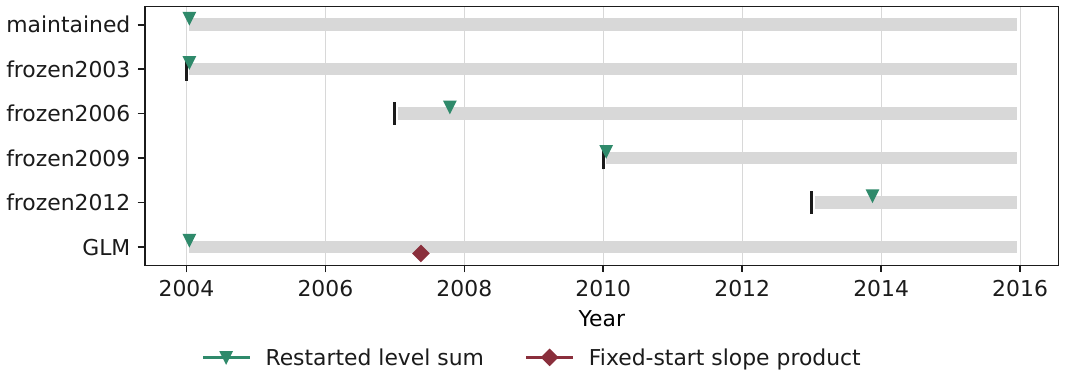}
\caption{First development alarms from the fixed-start slope product
and restarted level sum, both at threshold 20. The slope boundary
is 0.646. A vertical tick marks the month a model was frozen.}
\label{fig:paths}
\end{figure*}

At the common slope boundary of 0.646, the fixed-start slope product
crosses in May 2007 on the spline GLM and stays below 20 on the other
five development paths (Figure~\ref{fig:paths}).
The restarted level sum crosses on all six paths
(Table~\ref{tab:panel-alarms}).

The restart-intersection rule signals upward in January 2005 on maintained and
downward in March 2005 on frozen2003, March 2006 on the GLM, and August
2011 on frozen2009. The maintained alarm coincides with its first
January refit after monitoring begins. Its January-only set begins
at 0.765, above the 0.729 upper endpoint of the set started in August
2004. Under a constant slope of 0.759 on the same 144 monthly books,
the null alarm fractions are 6.15--6.25\% for maintained,
6.45--6.54\% for frozen2003 and 6.74\% for the GLM.
Table~\ref{tab:panel-calibration} gives all book sequences and slope pins.

Figure~\ref{fig:cs} shows terminal slopes of 0.875, 0.710 and 0.788
on maintained, frozen2003 and the GLM. The frozen2003 sequence
rejects a constant slope of 0.746, with log mixture evidence of 11.5.
Against its own commissioning boundary of 0.714, the GLM's
full-window fit of 0.788 is within tolerance.

\subsection{Chronological holdout results}\label{sec:holdout}

The restarted level sum starts in January 2016 and first crosses 20
in January 2018 on both paths. It exceeds 20 in 69 of 116 months on
maintained and 70 on frozen2015. The fixed-start slope product stays
below threshold on both paths (Table~\ref{tab:panel-alarms}).

For descriptive context, the cumulative slope fits at the end of the
holdout are 0.907 with band $[0.890,0.923]$ for maintained (January
2004--August 2025) and 0.760 with band $[0.740,0.778]$ for frozen2015
(January 2016--August 2025).

The restart-intersection rule signals downward on frozen2015 in
April 2018. In the supplementary maintained run started in January
2016, it signals upward in August 2019. The August-only set
starts at 0.932, above the 0.916 upper endpoint of the set started
in April 2017. Over these 116 monthly books, null alarm fractions
at slope 0.759 are 4.10\% and 1.56\% for maintained and frozen2015.
At slope one they are 3.32\% and 1.86--1.95\%
(Table~\ref{tab:panel-calibration}). The grid and continuous-set
implementations agree on the five development and frozen holdout alarm dates
(Table~\ref{tab:exact-intersections}).

\subsection{Simultaneous coverage and interval width}\label{sec:e12}

On the two development paths in Table~\ref{tab:battery}, the conditional
slope differs from monthly joint recalibration by at most 0.0014.

Panel A of Table~\ref{tab:coverage} compares intervals based on the
same cumulative conditional likelihood on 500 paired paths per setting,
with fixed books of 4{,}000 loans and constant slopes. The two unbounded
allocations spend
$\alpha/[t(t+1)]$ and $6\alpha/(\pi^2t^2)$ at look $t$.
A conditional Chernoff bound gives a two-sided log-likelihood drop of
$\log(2/\alpha_t)$ at each look, and summing the allocated errors
controls the probability of missing the true slope at any look.
Bonferroni uses $\alpha/36$ for 36 looks. The appendix tables also
give coverage at the asymptotic Wilks cutoff. Panel B gives monthly-refit
references at slope one and 36 looks.

In the 36-look comparisons both confidence sequences and the Chernoff
Bonferroni intervals give similar widths. At look 240 with 150 expected
events per month and slope 0.875, the continuous prior gives a median
width of 0.0439, compared with 0.0580 for the 29-node prior and 0.0605
and 0.0615 for the two unbounded spending rules. Their respective
every-look coverage rates are 98.4\%, 98.6\%, 99.4\% and 99.6\%.
At the on-node slope 0.70, the discrete mixture gives width 0.0396
and the continuous mixture gives 0.0430. Across all settings at
look 240, both mixtures have 97.0--99.6\% every-look coverage.
Both spending rules cover on 98.8--100.0\% of paths, and the continuous
sequence is 27.4--34.3\% narrower than the telescoping interval.

For a fixed discrete prior, a truth between nodes can have higher
expected log likelihood than every supported slope. As information
accumulates, the best supported node then sets a likelihood level
below the peak, leaving a positive width. At slope 0.875 in the
expected-likelihood stress, the 29-node width is 0.04995 after
240{,}000 equivalent looks. With a refined continuous uniform prior
the width is 0.00166, since the integral includes slopes arbitrarily
close to the peak.

\begin{table*}[t]
\centering
\caption{Coverage and interval width.
Each cell gives every-look/final-look coverage (\%), the two binomial
Monte Carlo standard errors (percentage points), and [median final
width]. Panel A uses 500 paths per setting at 150 expected events per
month. The truth is on the 29-node grid at 0.70 and off it at 0.875.
The continuous arm uses outer sets from a fixed 225-point mesh, with
a maximum log-integral gap of 0.647 nats at look 240. Bonferroni/36 ends
at look 36. Panel B uses a separate simulation of 800 paths at slope 1
and refits at each eligible look. Tables~\ref{tab:coverage-full}--\ref{tab:coverage-full-150}
report all slope and event-rate settings.}
\label{tab:coverage}
\footnotesize
\setlength{\tabcolsep}{6pt}
\begin{tabular}{@{}lcccc@{}}
\toprule
\multicolumn{5}{@{}l}{\textbf{A. Cumulative conditional likelihood}} \\
& \multicolumn{2}{c}{36 looks} & \multicolumn{2}{c}{240 looks} \\
\cmidrule(lr){2-3}\cmidrule(l){4-5}
Method & $0.70$ & $0.875$ & $0.70$ & $0.875$ \\
\midrule
CS, 29 nodes & 99.6/100.0 & 99.0/100.0 & 99.6/100.0 & 98.6/100.0 \\
 & (0.28/0.00) [0.1024] & (0.44/0.00) [0.1067] & (0.28/0.00) [0.0396] & (0.53/0.00) [0.0580] \\[2pt]
CS, continuous uniform & 99.6/100.0 & 98.8/99.8 & 99.4/100.0 & 98.4/100.0 \\
 & (0.28/0.00) [0.1033] & (0.49/0.20) [0.1054] & (0.35/0.00) [0.0430] & (0.56/0.00) [0.0439] \\[2pt]
Telescoping spend & 99.6/100.0 & 99.6/100.0 & 99.6/100.0 & 99.4/100.0 \\
 & (0.28/0.00) [0.1318] & (0.28/0.00) [0.1347] & (0.28/0.00) [0.0592] & (0.35/0.00) [0.0605] \\[2pt]
Inverse-square spend & 99.4/100.0 & 99.6/100.0 & 99.4/100.0 & 99.6/100.0 \\
 & (0.35/0.00) [0.1346] & (0.28/0.00) [0.1376] & (0.35/0.00) [0.0602] & (0.28/0.00) [0.0615] \\[2pt]
Chernoff Bonferroni/36 & 99.8/100.0 & 100.0/100.0 & --- & --- \\
 & (0.20/0.00) [0.1077] & (0.00/0.00) [0.1101] &  &  \\[2pt]
\bottomrule
\end{tabular}
\par\medskip
\begin{tabular}{@{}lccc@{}}
\toprule
\multicolumn{4}{@{}l}{\textbf{B. Monthly refits at 36 looks, true slope 1}} \\
& \multicolumn{3}{c}{Expected events per month} \\
\cmidrule(l){2-4}
Method & 10 & 40 & 150 \\
\midrule
Monthly profile & 15.88/93.88 & 13.75/95.25 & 14.88/96.00 \\
 & (1.29/0.85) [1.20] & (1.22/0.75) [0.63] & (1.26/0.69) [0.34] \\[2pt]
Monthly profile Bonferroni/36 & 94.25/99.62 & 95.50/100.00 & 94.38/99.88 \\
 & (0.82/0.22) [1.82] & (0.73/0.00) [1.04] & (0.81/0.12) [0.57] \\[2pt]
Monthly Wald & 18.25/94.62 & 15.00/95.25 & 15.25/96.12 \\
 & (1.37/0.80) [1.27] & (1.26/0.75) [0.64] & (1.27/0.68) [0.35] \\[2pt]
\bottomrule
\end{tabular}
\end{table*}

Appendix~\ref{app:secondary} reports vintage-level recalibration and
simulations of monitoring duration by book size.

\section{Robustness analyses}\label{sec:limits}

\textbf{Restarted slope sum.}
Each cutoff in Table~\ref{tab:panel-calibration} is retrospective,
calibrated over the run's realized books and counts. At those cutoffs,
frozen2003 and the GLM cross, and frozen2006 stays below its cutoff.
Appendices~\ref{app:secondary} and~\ref{app:panel-calibration} report
the synthetic comparisons and actual-book calibration, respectively.

\textbf{Curvature.}
The curvature-profiled slope product stays below 20 on every path,
including the GLM at the common boundary of 0.646. It tests the local
slope at a fixed score under a quadratic conditional model. Median
absolute curvature ranges from 0.025 to 0.085 across the eight paths.
Appendix~\ref{sec:curvature} gives the model adequacy checks, curvature
simulations and the sensitivity to common and group-level shocks.

\section{Conclusion}

Conditioning on monthly event counts lets a validation team test a
slope tolerance without estimating a path of calibration intercepts.
On the mortgage panel, the fixed-slope benchmark reproduces 78\% of
maintained AUC variance in development and 70\% in the holdout.
The fixed-start slope product at 20 bounds the chance of ever
crossing by 5\% under the conditional-logit null. In the holdout, the
restarted level sum crossed on both paths and the fixed-start slope
product stayed below threshold on both. The restart-intersection rule
signaled a downward change on the frozen path and an upward change
in the supplementary maintained run.

\bibliographystyle{ACM-Reference-Format}
\bibliography{refs}

\appendix

\section{Calibration level and the rare-event limit of AUC}\label{app:auc}

Under \eqref{eq:model}, AUC compares the score of an event with that of
a non-event, with half credit for ties. The two scores are drawn
from distributions proportional to $f(\theta)\sig(\alpha+\beta\theta)$
and $f(\theta)\{1-\sig(\alpha+\beta\theta)\}$, respectively, where
$f$ is the book's score distribution. As $\alpha\to-\infty$,
$\sig(\alpha+\beta\theta)\sim e^{\alpha}e^{\beta\theta}$.
The common factor $e^{\alpha}$ cancels in the normalized event-score
distribution in that limit. At finite $\alpha$ both class-conditional
distributions can depend on the intercept. In the exploratory
200{,}000-loan calculation at $\beta=1$, varying $\alpha$ from $-1.22$
to $+0.51$ changes attainable AUC by 0.0034.

\section{Numerical implementation}\label{app:numerics}

The 29-node prior puts equal weight on nodes 0.05 apart.
Convexity of $\log Z_{M_t}$ makes $A_T$ concave and
$\log\mathcal M_T(\beta_0)$ convex. The mixture ratio at a maximizing node
is at most one, so the acceptance interval contains a node. The outer
neighbors of the included nodes enclose it within the reporting range.
We interpolate the log-normalizer on 561 points for display. Direct
checks place the displayed terminal development endpoints inside the
exact sets. An endpoint at the reporting limit marks truncation.

The fixed-start slope product and restarted slope sum average
likelihood ratios at offsets 0.05, 0.10, 0.20, 0.35 and 0.50 below
the boundary. The grid implementation of the restart-intersection rule
evaluates the interpolated likelihood on the reporting grid.
For the continuous-set recheck we first evaluate each conditional
likelihood on 1{,}009 nodes, identify the restarts with the binding
endpoint brackets, and bracket the continuous endpoints by direct
likelihood evaluation (Table~\ref{tab:exact-intersections}).
A retrospective direction diagnostic averages
cumulative slope fits before and from the alarm month onward. On the
maintained development path these averages are 0.765 and 0.816.
On the maintained holdout reset they are 0.954 and 0.927. The
retrospective decrease on the reset differs from the upward separation
of the binding sets in August 2019.

For Table~\ref{tab:battery}, the relative Monte Carlo standard error of
the development null SD is 0.74\%. Propagating this through the squared
SD gives about $1.47(1-\rho)$ percentage points for the raw excess share,
or 1.15 points at $\rho=22\%$.

\section{Proofs of the sequential guarantees}\label{app:proofs}

\begin{proof}[Proof of Proposition~\ref{prop:eprocess}]
Write $\mathcal H_t=\mathcal F_{t-1}\vee\sigma(\alpha_t)$ and
$\mathcal G_t=\mathcal H_t\vee\sigma(M_t)$. Let $S_t$ be the event set.
Given $\mathcal H_t$, independence and the Bernoulli odds in
\eqref{eq:model} give, for a subset $A$ of size $m$,
\[
\Pb(S_t=A\mid\mathcal H_t)
=\frac{\exp\{m\alpha_t+\beta_t\sum_{i\in A}\theta_{it}\}}
       {\prod_i(1+\exp\{\alpha_t+\beta_t\theta_{it}\})}.
\]
Dividing by the sum of this probability over subsets of size $m$
cancels the common intercept and denominator. Thus given $\mathcal G_t$,
$S_t$ has the law in \eqref{eq:arrangement} at $b=\beta_t$ for each
realization of the intercept.

Fix that history and the count. Write $\Eb_b$ for expectation under
the resulting finite conditional family at slope $b$. For fixed
$\beta_1<\beta_0$, the likelihood ratio is a decreasing function
$g(T_t)=E_t(\beta_1,\beta_0)$ of the sufficient statistic. Summing the
likelihood ratio against its denominator law gives
$\Eb_{\beta_0}[g(T_t)]=1$. Differentiating the finite exponential-family
sum gives
\[
\frac{d}{db}\Eb_b[g(T_t)]
  =\operatorname{Cov}_b(g(T_t),T_t)\le 0.
\]
Indeed, twice the covariance equals
$\Eb_b[(g(T_t)-g(T'_t))(T_t-T'_t)]$ for an independent copy $T'_t$,
and the integrand is nonpositive. This includes degenerate counts,
for which the ratio is one. We also set the factor to one when $M_t<5$,
a choice measurable in $\mathcal G_t$. For $b=\beta_t\ge\beta_0$ it follows that
$\Eb[E_t\mid\mathcal G_t]\le1$. A predictable probability mixture over
$\beta_1<\beta_0$ preserves the bound by conditional integration.

Averaging first over the intercept gives the bound conditional on
$\mathcal F_{t-1},M_t$. Averaging over the count gives
$\Eb[E_t\mid\mathcal F_{t-1}]\le1$. Since $R_{t-1}$ is nonnegative and
$\mathcal F_{t-1}$-measurable,
$\Eb[R_t\mid\mathcal F_{t-1}]\le R_{t-1}$. Starting from $R_0=1$,
induction gives integrability and the nonnegative supermartingale
property. Ville's inequality \cite{ville1939} gives
$\Pb(\sup_t R_t\ge 1/\alpha)\le\alpha$.
\end{proof}

\begin{proof}[Proof of Proposition~\ref{prop:cs}]
Under $\beta_t=\beta^\star$, the conditional distribution given
$\mathcal G_t$ is the denominator law for every ratio
$E_t(u,\beta^\star)$. Its conditional mean is one for any $u$ in the
slope domain because numerator and denominator have the same support
and the numerator probabilities sum to one. The tower property gives
mean one conditional on $\mathcal F_{t-1}$. Thus for a start $r$, each
product $\prod_{s=r}^t E_s(u,\beta^\star)$ is a nonnegative
martingale initialized at one at $r-1$.

At any candidate $b$ define the restarted likelihood-ratio mixture
\[
\mathcal M_{r,t}(b)=
\int\exp\{A_t(u)-A_{r-1}(u)-A_t(b)+A_{r-1}(b)\}\,\Pi(du).
\]
At $b=\beta^\star$ conditional integration preserves the martingale
property by nonnegativity. Taking $r=1$ gives the mean-one process in
the proposition.
For every start $r$, Ville's inequality conditional on
$\mathcal F_{r-1}$ bounds the probability that
$\mathcal M_{r,t}(\beta^\star)$ ever reaches $1/\alpha$ by $\alpha$.
Excluding $b$ at time $t$ means $\mathcal M_{r,t}(b)\ge 1/\alpha$,
which gives time-uniform coverage for each fixed start $r$.
\end{proof}

\paragraph{Run length of the restart-intersection rule.}
Let $\Theta$ be the slope domain, an interval containing the constant
true slope $\beta^\star$, and use the exact sets
$\mathrm{CS}_{r,t}=\{b\in\Theta:\mathcal M_{r,t}(b)<1/\alpha\}$.
This is the forward restart construction of \citet{shekhar2024}.
For the current-set intersection used here the bound follows directly
from the restarted mixtures. Under no change and the stated monthly
model, $L_t=\sum_{r=1}^t\mathcal M_{r,t}(\beta^\star)$ satisfies
$\Eb[L_t\mid\mathcal F_{t-1}]=L_{t-1}+1$ with $L_0=0$.
At an empty intersection, some set excludes $\beta^\star$, so
$L_\tau\ge1/\alpha$. The exact sets are relatively open in $\Theta$
by continuity of the likelihood ratios. A countable dense subset of
$\Theta$ determines whether the intersection is empty, so $\tau$ is a
stopping time. Bounded optional stopping for $L_t-t$ gives
\[
\Eb[\tau\wedge n]=\Eb[L_{\tau\wedge n}]
  \ge \alpha^{-1}\Pb(\tau\le n).
\]
If $\tau$ is finite almost surely, taking $n\to\infty$ gives
$\Eb[\tau]\ge1/\alpha$ by monotone convergence. If $\tau=\infty$ with
positive probability, the mean is infinite.

\section{Supplementary comparisons and study outcomes}\label{app:hyps}

At 150 expected events per month, Bonferroni/36 with the asymptotic
Wilks cutoff gives 99.0\% every-look and 100.0\% final-look coverage
at slope 0.70 (Monte Carlo SEs 0.44 and 0.00 percentage points).
At slope 0.875, the corresponding coverage is 98.0\% and 99.8\%
(SEs 0.63 and 0.20 percentage points).
Table~\ref{tab:hyps} records the preregistered criteria and outcomes.

\begin{table*}[t]
\centering
\caption{Targets and scope of sequential monitoring methods.
Intercept invariance refers to elimination of a logistic intercept while
retaining the monitored parameter. A dash marks a feature outside the
method's scope.}
\label{tab:related-comparison}
\footnotesize
\setlength{\tabcolsep}{4pt}
\begin{tabularx}{\linewidth}{@{}>{\raggedright\arraybackslash}p{.22\linewidth}
  >{\raggedright\arraybackslash}X c
  >{\raggedright\arraybackslash}p{.12\linewidth}
  >{\raggedright\arraybackslash}X >{\raggedright\arraybackslash}X@{}}
\toprule
Method & Target & \shortstack{Intercept\\invariant} & Anytime-valid
 & Change-point rule & Real financial application \\
\midrule
This paper & Calibration slope & Yes & Yes & Restart-intersection rule & Mortgage panel \\
\citet{arnold2023} & PIT calibration & --- & Yes & --- & --- \\
\citet{wangEbacktesting2026} & Risk measures & --- & Yes & --- & Market data \\
\citet{farran2026} (PITMonitor) & PIT distribution & --- & Yes & Mixture alarm and estimation & --- \\
\citet{henzi2023} & Binary probability calibration & --- & Yes & --- & Credit application \\
\citet{hao2024} (COND) & Exponential-family parameter &
Yes in this model & Fixed-sample factor & --- & --- \\
\bottomrule
\end{tabularx}
\end{table*}

\begin{table*}[t]
\centering
\caption{10 expected events per month. Every-look coverage (\%) (binomial Monte Carlo standard
error in percentage points) [median final width] at each slope and
event rate. Each setting has 500 paths. The tested slopes on the
29-node grid are 0.70 and 1.00. In order, the last four methods are
telescoping and inverse-square Chernoff spending, Chernoff
Bonferroni/36, and Wilks Bonferroni/36.}
\label{tab:coverage-full}
\footnotesize
\setlength{\tabcolsep}{2pt}
\begin{tabular}{@{}lcccccc@{}}
\toprule
Slope & CS 29 & CS uniform & Telescoping & Inv. square & Bonf./36 & Wilks/36 \\
\midrule
\multicolumn{7}{@{}l}{36 looks} \\
0.646 & \coverageentry{97.8}{0.66}{0.3415} & \coverageentry{97.8}{0.66}{0.3404} & \coverageentry{99.4}{0.35}{0.4845} & \coverageentry{99.4}{0.35}{0.4949} & \coverageentry{99.8}{0.20}{0.3961} & \coverageentry{98.4}{0.56}{0.3320} \\
0.7 & \coverageentry{98.6}{0.53}{0.3419} & \coverageentry{98.6}{0.53}{0.3408} & \coverageentry{99.4}{0.35}{0.4851} & \coverageentry{99.2}{0.40}{0.4955} & \coverageentry{100.0}{0.00}{0.3966} & \coverageentry{99.4}{0.35}{0.3324} \\
0.875 & \coverageentry{97.6}{0.68}{0.3401} & \coverageentry{97.6}{0.68}{0.3390} & \coverageentry{99.6}{0.28}{0.4824} & \coverageentry{99.4}{0.35}{0.4927} & \coverageentry{99.8}{0.20}{0.3943} & \coverageentry{98.8}{0.49}{0.3305} \\
0.93 & \coverageentry{98.4}{0.56}{0.3403} & \coverageentry{98.2}{0.59}{0.3392} & \coverageentry{100.0}{0.00}{0.4828} & \coverageentry{99.8}{0.20}{0.4931} & \coverageentry{100.0}{0.00}{0.3946} & \coverageentry{98.6}{0.53}{0.3307} \\
1 & \coverageentry{97.8}{0.66}{0.3389} & \coverageentry{97.8}{0.66}{0.3379} & \coverageentry{100.0}{0.00}{0.4806} & \coverageentry{100.0}{0.00}{0.4909} & \coverageentry{99.8}{0.20}{0.3928} & \coverageentry{98.4}{0.56}{0.3292} \\
1.07 & \coverageentry{98.2}{0.59}{0.3388} & \coverageentry{98.2}{0.59}{0.3377} & \coverageentry{99.2}{0.40}{0.4805} & \coverageentry{99.0}{0.44}{0.4908} & \coverageentry{99.8}{0.20}{0.3927} & \coverageentry{98.6}{0.53}{0.3291} \\
1.33 & \coverageentry{98.6}{0.53}{0.3409} & \coverageentry{98.6}{0.53}{0.3399} & \coverageentry{100.0}{0.00}{0.4798} & \coverageentry{100.0}{0.00}{0.4885} & \coverageentry{100.0}{0.00}{0.3948} & \coverageentry{98.8}{0.49}{0.3316} \\
\midrule
\multicolumn{7}{@{}l}{240 looks} \\
0.646 & \coverageentry{97.6}{0.68}{0.1437} & \coverageentry{97.6}{0.68}{0.1434} & \coverageentry{99.4}{0.35}{0.2182} & \coverageentry{99.4}{0.35}{0.2218} & --- & --- \\
0.7 & \coverageentry{98.2}{0.59}{0.1438} & \coverageentry{98.2}{0.59}{0.1434} & \coverageentry{99.4}{0.35}{0.2183} & \coverageentry{99.2}{0.40}{0.2220} & --- & --- \\
0.875 & \coverageentry{97.0}{0.76}{0.1427} & \coverageentry{97.0}{0.76}{0.1424} & \coverageentry{99.6}{0.28}{0.2167} & \coverageentry{99.4}{0.35}{0.2203} & --- & --- \\
0.93 & \coverageentry{97.8}{0.66}{0.1428} & \coverageentry{97.4}{0.71}{0.1424} & \coverageentry{100.0}{0.00}{0.2166} & \coverageentry{99.8}{0.20}{0.2202} & --- & --- \\
1 & \coverageentry{97.6}{0.68}{0.1424} & \coverageentry{97.6}{0.68}{0.1422} & \coverageentry{100.0}{0.00}{0.2162} & \coverageentry{100.0}{0.00}{0.2199} & --- & --- \\
1.07 & \coverageentry{97.8}{0.66}{0.1425} & \coverageentry{97.8}{0.66}{0.1421} & \coverageentry{99.2}{0.40}{0.2161} & \coverageentry{99.0}{0.44}{0.2197} & --- & --- \\
1.33 & \coverageentry{97.6}{0.68}{0.1434} & \coverageentry{97.6}{0.68}{0.1429} & \coverageentry{100.0}{0.00}{0.2174} & \coverageentry{100.0}{0.00}{0.2210} & --- & --- \\
\bottomrule
\end{tabular}
\end{table*}

\begin{table*}[t]
\centering
\caption{40 expected events per month. Every-look coverage (\%) (binomial Monte Carlo standard
error in percentage points) [median final width] at each slope and
event rate. Each setting has 500 paths. The tested slopes on the
29-node grid are 0.70 and 1.00. In order, the last four methods are
telescoping and inverse-square Chernoff spending, Chernoff
Bonferroni/36, and Wilks Bonferroni/36.}
\label{tab:coverage-full-40}
\footnotesize
\setlength{\tabcolsep}{2pt}
\begin{tabular}{@{}lcccccc@{}}
\toprule
Slope & CS 29 & CS uniform & Telescoping & Inv. square & Bonf./36 & Wilks/36 \\
\midrule
\multicolumn{7}{@{}l}{36 looks} \\
0.646 & \coverageentry{98.0}{0.63}{0.1828} & \coverageentry{98.0}{0.63}{0.1823} & \coverageentry{99.4}{0.35}{0.2443} & \coverageentry{99.4}{0.35}{0.2495} & \coverageentry{100.0}{0.00}{0.1997} & \coverageentry{98.8}{0.49}{0.1674} \\
0.7 & \coverageentry{99.6}{0.28}{0.1828} & \coverageentry{99.6}{0.28}{0.1823} & \coverageentry{100.0}{0.00}{0.2443} & \coverageentry{100.0}{0.00}{0.2495} & \coverageentry{100.0}{0.00}{0.1997} & \coverageentry{99.2}{0.40}{0.1674} \\
0.875 & \coverageentry{99.0}{0.44}{0.1835} & \coverageentry{99.0}{0.44}{0.1830} & \coverageentry{99.8}{0.20}{0.2453} & \coverageentry{99.8}{0.20}{0.2506} & \coverageentry{100.0}{0.00}{0.2005} & \coverageentry{98.8}{0.49}{0.1681} \\
0.93 & \coverageentry{98.2}{0.59}{0.1836} & \coverageentry{98.2}{0.59}{0.1832} & \coverageentry{99.2}{0.40}{0.2456} & \coverageentry{99.2}{0.40}{0.2509} & \coverageentry{100.0}{0.00}{0.2007} & \coverageentry{98.2}{0.59}{0.1683} \\
1 & \coverageentry{98.4}{0.56}{0.1842} & \coverageentry{98.4}{0.56}{0.1838} & \coverageentry{99.4}{0.35}{0.2465} & \coverageentry{99.0}{0.44}{0.2518} & \coverageentry{99.8}{0.20}{0.2015} & \coverageentry{98.4}{0.56}{0.1689} \\
1.07 & \coverageentry{98.6}{0.53}{0.1848} & \coverageentry{98.6}{0.53}{0.1844} & \coverageentry{99.0}{0.44}{0.2474} & \coverageentry{98.8}{0.49}{0.2527} & \coverageentry{99.8}{0.20}{0.2022} & \coverageentry{99.0}{0.44}{0.1695} \\
1.33 & \coverageentry{99.0}{0.44}{0.1906} & \coverageentry{99.0}{0.44}{0.1901} & \coverageentry{99.8}{0.20}{0.2558} & \coverageentry{99.8}{0.20}{0.2612} & \coverageentry{99.6}{0.28}{0.2090} & \coverageentry{98.0}{0.63}{0.1752} \\
\midrule
\multicolumn{7}{@{}l}{240 looks} \\
0.646 & \coverageentry{97.8}{0.66}{0.0741} & \coverageentry{97.8}{0.66}{0.0761} & \coverageentry{99.4}{0.35}{0.1098} & \coverageentry{99.4}{0.35}{0.1117} & --- & --- \\
0.7 & \coverageentry{99.4}{0.35}{0.0737} & \coverageentry{99.4}{0.35}{0.0761} & \coverageentry{100.0}{0.00}{0.1099} & \coverageentry{100.0}{0.00}{0.1117} & --- & --- \\
0.875 & \coverageentry{99.0}{0.44}{0.0814} & \coverageentry{98.8}{0.49}{0.0763} & \coverageentry{99.8}{0.20}{0.1101} & \coverageentry{99.8}{0.20}{0.1120} & --- & --- \\
0.93 & \coverageentry{98.0}{0.63}{0.0806} & \coverageentry{97.8}{0.66}{0.0764} & \coverageentry{99.2}{0.40}{0.1103} & \coverageentry{99.2}{0.40}{0.1121} & --- & --- \\
1 & \coverageentry{98.2}{0.59}{0.0745} & \coverageentry{98.2}{0.59}{0.0767} & \coverageentry{99.4}{0.35}{0.1107} & \coverageentry{99.0}{0.44}{0.1126} & --- & --- \\
1.07 & \coverageentry{98.4}{0.56}{0.0806} & \coverageentry{98.2}{0.59}{0.0770} & \coverageentry{99.0}{0.44}{0.1113} & \coverageentry{98.8}{0.49}{0.1131} & --- & --- \\
1.33 & \coverageentry{99.0}{0.44}{0.0832} & \coverageentry{99.0}{0.44}{0.0793} & \coverageentry{99.8}{0.20}{0.1148} & \coverageentry{99.8}{0.20}{0.1168} & --- & --- \\
\bottomrule
\end{tabular}
\end{table*}

\begin{table*}[t]
\centering
\caption{150 expected events per month. Every-look coverage (\%) (binomial Monte Carlo standard
error in percentage points) [median final width] at each slope and
event rate. Each setting has 500 paths. The tested slopes on the
29-node grid are 0.70 and 1.00. In order, the last four methods are
telescoping and inverse-square Chernoff spending, Chernoff
Bonferroni/36, and Wilks Bonferroni/36.}
\label{tab:coverage-full-150}
\footnotesize
\setlength{\tabcolsep}{2pt}
\begin{tabular}{@{}lcccccc@{}}
\toprule
Slope & CS 29 & CS uniform & Telescoping & Inv. square & Bonf./36 & Wilks/36 \\
\midrule
\multicolumn{7}{@{}l}{36 looks} \\
0.646 & \coverageentry{99.6}{0.28}{0.1020} & \coverageentry{99.8}{0.20}{0.1029} & \coverageentry{99.6}{0.28}{0.1312} & \coverageentry{99.6}{0.28}{0.1340} & \coverageentry{100.0}{0.00}{0.1072} & \coverageentry{99.0}{0.44}{0.0899} \\
0.7 & \coverageentry{99.6}{0.28}{0.1024} & \coverageentry{99.6}{0.28}{0.1033} & \coverageentry{99.6}{0.28}{0.1318} & \coverageentry{99.4}{0.35}{0.1346} & \coverageentry{99.8}{0.20}{0.1077} & \coverageentry{99.0}{0.44}{0.0903} \\
0.875 & \coverageentry{99.0}{0.44}{0.1067} & \coverageentry{98.8}{0.49}{0.1054} & \coverageentry{99.6}{0.28}{0.1347} & \coverageentry{99.6}{0.28}{0.1376} & \coverageentry{100.0}{0.00}{0.1101} & \coverageentry{98.0}{0.63}{0.0923} \\
0.93 & \coverageentry{98.6}{0.53}{0.1075} & \coverageentry{98.8}{0.49}{0.1063} & \coverageentry{99.4}{0.35}{0.1360} & \coverageentry{99.0}{0.44}{0.1389} & \coverageentry{99.8}{0.20}{0.1111} & \coverageentry{98.2}{0.59}{0.0932} \\
1 & \coverageentry{98.6}{0.53}{0.1072} & \coverageentry{98.6}{0.53}{0.1076} & \coverageentry{99.2}{0.40}{0.1378} & \coverageentry{99.2}{0.40}{0.1407} & \coverageentry{99.6}{0.28}{0.1126} & \coverageentry{97.6}{0.68}{0.0944} \\
1.07 & \coverageentry{98.6}{0.53}{0.1102} & \coverageentry{98.6}{0.53}{0.1092} & \coverageentry{98.8}{0.49}{0.1399} & \coverageentry{98.8}{0.49}{0.1429} & \coverageentry{99.8}{0.20}{0.1144} & \coverageentry{98.4}{0.56}{0.0959} \\
1.33 & \coverageentry{99.2}{0.40}{0.1173} & \coverageentry{99.2}{0.40}{0.1166} & \coverageentry{99.4}{0.35}{0.1502} & \coverageentry{99.4}{0.35}{0.1534} & \coverageentry{99.8}{0.20}{0.1228} & \coverageentry{99.0}{0.44}{0.1029} \\
\midrule
\multicolumn{7}{@{}l}{240 looks} \\
0.646 & \coverageentry{99.6}{0.28}{0.0402} & \coverageentry{99.6}{0.28}{0.0428} & \coverageentry{99.6}{0.28}{0.0589} & \coverageentry{99.6}{0.28}{0.0599} & --- & --- \\
0.7 & \coverageentry{99.6}{0.28}{0.0396} & \coverageentry{99.4}{0.35}{0.0430} & \coverageentry{99.6}{0.28}{0.0592} & \coverageentry{99.4}{0.35}{0.0602} & --- & --- \\
0.875 & \coverageentry{98.6}{0.53}{0.0580} & \coverageentry{98.4}{0.56}{0.0439} & \coverageentry{99.4}{0.35}{0.0605} & \coverageentry{99.6}{0.28}{0.0615} & --- & --- \\
0.93 & \coverageentry{98.0}{0.63}{0.0562} & \coverageentry{98.2}{0.59}{0.0443} & \coverageentry{99.4}{0.35}{0.0611} & \coverageentry{99.0}{0.44}{0.0621} & --- & --- \\
1 & \coverageentry{98.4}{0.56}{0.0415} & \coverageentry{98.4}{0.56}{0.0448} & \coverageentry{99.2}{0.40}{0.0619} & \coverageentry{99.2}{0.40}{0.0629} & --- & --- \\
1.07 & \coverageentry{98.4}{0.56}{0.0571} & \coverageentry{98.2}{0.59}{0.0454} & \coverageentry{98.8}{0.49}{0.0629} & \coverageentry{98.8}{0.49}{0.0639} & --- & --- \\
1.33 & \coverageentry{99.2}{0.40}{0.0586} & \coverageentry{99.2}{0.40}{0.0485} & \coverageentry{99.4}{0.35}{0.0675} & \coverageentry{99.4}{0.35}{0.0686} & --- & --- \\
\bottomrule
\end{tabular}
\end{table*}

\begin{table*}[t]
\centering
\caption{First alarms from the grid and continuous-set implementations.
Signed width is the smallest upper endpoint minus the largest lower endpoint
at the alarm. Direction is read from the newer binding set. The maintained
holdout run is supplementary.}
\label{tab:exact-intersections}
\footnotesize
\setlength{\tabcolsep}{3pt}
\begin{tabular}{@{}lrrrlr@{}}
\toprule
Path & Run start & Grid & Continuous & Direction & Signed width \\
\midrule
maintained & 2004-01 & 2005-01 & 2005-01 & Up & $-0.03537$ \\
frozen2003 & 2004-01 & 2005-03 & 2005-03 & Down & $-0.02176$ \\
GLM & 2004-01 & 2006-03 & 2006-03 & Down & $-0.06845$ \\
frozen2009 & 2010-01 & 2011-08 & 2011-08 & Down & $-0.02649$ \\
frozen2015 & 2016-01 & 2018-04 & 2018-04 & Down & $-0.06156$ \\
maintained & 2016-01 & --- & 2019-08 & Up & $-0.01658$ \\
\bottomrule
\end{tabular}
\end{table*}

\begin{table*}[t]
\centering
\caption{Conditional model adequacy and one-sided alarms. LR rejections count
months with nominal $p<0.05$. Dates refer to the restarted slope sum
at threshold 20. The lagged column treats the previous 12-month median
curvature as known. Binned slopes use monthly score deciles.}
\label{tab:conditional-adequacy}
\footnotesize
\setlength{\tabcolsep}{3pt}
\begin{tabular}{@{}lrrrrlllrrr@{}}
\toprule
& & & \multicolumn{2}{c}{LR rejections} & \multicolumn{3}{c}{First alarm} & \multicolumn{3}{c}{Binned slope} \\
\cmidrule(lr){4-5}\cmidrule(lr){6-8}\cmidrule(l){9-11}
Path & Months & Median $|\gamma|$ & Quad. & Spline & Linear & Lagged & Profiled & Stated & Linear & Quad. \\
\midrule
\multicolumn{11}{@{}l}{Development} \\
maintained & 144 & 0.031 & 16 & 22 & --- & --- & --- & 0.890 & 1.023 & 1.028 \\
frozen2003 & 144 & 0.071 & 47 & 46 & 2005-03 & 2005-03 & 2006-10 & 0.665 & 0.921 & 1.006 \\
frozen2006 & 108 & 0.085 & 40 & 37 & 2011-08 & 2009-08 & 2011-08 & 0.802 & 0.940 & 1.023 \\
frozen2009 & 72 & 0.045 & 14 & 17 & --- & --- & --- & 0.969 & 1.030 & 1.064 \\
frozen2012 & 36 & 0.037 & 11 & 10 & --- & --- & --- & 0.896 & 0.990 & 1.050 \\
GLM & 144 & 0.083 & 40 & 39 & 2005-12 & 2006-03 & 2007-05 & 0.767 & 1.015 & 1.002 \\
\midrule
\multicolumn{11}{@{}l}{Holdout} \\
maintained & 116 & 0.034 & 19 & 21 & --- & --- & --- & 0.951 & 1.012 & 1.017 \\
frozen2015 & 116 & 0.025 & 21 & 17 & --- & 2024-04 & --- & 0.773 & 1.006 & 0.995 \\
\bottomrule
\end{tabular}
\end{table*}

\begin{table*}[t]
\centering
\caption{Quadratic-logit simulation over 120 looks on 500 paths
per setting with 40 expected events per look. Drift measures the shift
in mean score in book standard deviations. Rates are percentages
(Monte Carlo standard errors in percentage points). Coverage at every
look targets either that prefix's projection or the fixed terminal
projection.}
\label{tab:curvature-simulation}
\footnotesize
\setlength{\tabcolsep}{4pt}
\begin{tabular}{@{}lcrrrrr@{}}
\toprule
$\beta$ & $\gamma$ & Drift & Final projection & Alarm & Prefix coverage & Terminal coverage \\
\midrule
0.646 & 0 & $0$ & 0.6460 & 4.2 (0.90) & 98.8 (0.49) & 98.8 (0.49) \\
0.646 & 0 & $-0.5$ & 0.6460 & 4.2 (0.90) & 98.8 (0.49) & 98.8 (0.49) \\
0.646 & 0.1 & $0$ & 0.7878 & 0.2 (0.20) & 95.8 (0.90) & 95.8 (0.90) \\
0.646 & 0.1 & $-0.5$ & 0.7271 & 2.2 (0.66) & 95.6 (0.92) & 79.2 (1.82) \\
0.646 & 0.3 & $0$ & 1.1810 & 0.0 (0.00) & 88.6 (1.42) & 88.6 (1.42) \\
0.646 & 0.3 & $-0.5$ & 0.9212 & 7.8 (1.20) & 88.2 (1.44) & 0.0 (0.00) \\
0.700 & 0 & $0$ & 0.7000 & 0.0 (0.00) & 98.6 (0.53) & 98.6 (0.53) \\
0.700 & 0 & $-0.5$ & 0.7000 & 0.0 (0.00) & 98.6 (0.53) & 98.6 (0.53) \\
0.700 & 0.1 & $0$ & 0.8519 & 0.0 (0.00) & 94.8 (0.99) & 94.8 (0.99) \\
0.700 & 0.1 & $-0.5$ & 0.7915 & 0.6 (0.35) & 94.6 (1.01) & 80.6 (1.77) \\
0.700 & 0.3 & $0$ & 1.2669 & 0.0 (0.00) & 90.0 (1.34) & 90.0 (1.34) \\
0.700 & 0.3 & $-0.5$ & 1.0130 & 1.0 (0.44) & 87.6 (1.47) & 0.2 (0.20) \\
\bottomrule
\end{tabular}
\end{table*}

\begin{table*}[t]
\caption{Preregistered criteria and outcomes in the development and
holdout windows.}
\label{tab:hyps}
\small
\begin{tabular}{@{}p{0.30\textwidth}p{0.64\textwidth}@{}}
\toprule
hypothesis & outcome\\
\midrule
H1: $C^M$ explains most of realized-AUC variance ($R^2\ge0.7$; kill
$<0.3$) & \textbf{Not met}: $R^2$ 0.068--0.450 across paths
($<0.3$ on five of six).\\
H2: no slope alarm on maintained; slope alarm on frozen &
\textbf{Met} in development with the preregistered restarted slope
sum at 20:
no maintained alarm in 144 months (peak
log evidence 1.8 against 3.0); frozen2003 alarm in March 2005.
\textbf{Not met} in the holdout: maintained met its no-alarm criterion
(peak $-2.7$), while frozen2015 also stayed below threshold (peak 1.41).
Its restart-intersection rule signaled in April 2018. The fixed-start
slope product was designated the primary one-sided statistic after
the panel outcomes were available.\\
H3(i): no alarms on pure-level (N1) and pure-case-mix (N2) negatives where
level-sensitive baselines fire &
\textbf{Met on separation} in the preregistered evaluation. Under boundary
calibration the restarted slope sum has no alarms under N1 or N2 and the
restarted level sum alarms on every N1 path.\\
H3(ii): median detection delay $\le$ half the best baseline's &
\textbf{Not met} in the preregistered evaluation. For the break below
tolerance to 0.55 under boundary calibration, the restarted slope sum's
median delay is 45 months and the score CUSUM's is 16.\\
H5(a): rare-event approximation $\Phi(\sigma/\sqrt{2})$ for score
dispersion $\sigma$ tracks $C^M$ within $\pm0.02$ &
\textbf{Partial}: correlation 0.49--0.92; bias $+0.010$ to $+0.034$.\\
H5(b): rate-gap dispersion explains score dispersion ($R^2\ge0.6$) &
\textbf{Met} on the maintained path ($R^2=0.678$).\\
H6: alarm months show materially worse top-decile prepayment, 12-month
cash-flow and weighted-average-life errors at matched level error &
\textbf{Small increment}: adding slope evidence to level error raises
$R^2$ by at most 0.021 for top-decile prepayment error. We did not compute
the cash-flow or weighted-average-life endpoints.\\
\bottomrule
\end{tabular}
\end{table*}

\section{Secondary analyses}\label{app:secondary}

The restarted slope sum of products \cite{shin2024} is
$D_t=(1+D_{t-1})E_t$ with $D_0=0$. Under the null in
Proposition~\ref{prop:eprocess}, $D_t-t$ is a supermartingale,
and the mean time to false alarm at threshold 20 is at least 20 months.
For horizon-wide comparisons we calibrate the threshold by resampling.

\subsection{Boundary-calibrated monitor comparison}\label{sec:bakeoff}

We simulate monthly books of 4{,}000 loans with cyclical score
dispersion. Each change begins at look 49. P1 reduces the slope from
one over 12 months at the same mean event count. P2 permutes a share of
scores while generating outcomes from the unpermuted scores. N1 raises
the intercept over 24 months and N2 compresses score dispersion with
the conditional response unchanged. Calibration uses 400 separate
paths per book set and the book the monitor sees, with randomized ties.
PSI is constant on a fixed book sequence. Its matched alarm rate
reflects the randomized tie-break.

The restarted slope sum with boundary calibration alarms on 2.50\%
of paths at slope 0.646 and on none at 0.70 (Table~\ref{tab:bakeoff}). Calibrating
at the interior slope 0.70 gives 4.25\% at that slope and 23.25\% at
the boundary. At the industry thresholds the AUC chart alarms on
99.75\% and 100\% of paths at the two null slopes, respectively,
and PSI alarms on all paths at both slopes. The log threshold for
the restarted slope sum on the base synthetic books is 6.62.

At the boundary-calibrated thresholds the restarted slope sum detects the
break to 0.55 on 99\% of paths with a median delay of 45 months.
The score CUSUM detects it on 100\% with a median delay of 16 months
and alarms on 81\% under the larger intercept shift. The restarted
slope sum stays below threshold for drops to 0.85 and 0.70, both
within tolerance. Its alarm rate
is zero under both pure-level shifts and the case-mix control.
The restarted level sum alarms on all paths under both pure-level shifts.

The sharp-null ratio references an AUC distribution that fixes the
intercept. In the separate book-size sensitivity run, increasing the
intercept by 1.2 shifts its AUC center down by 0.0060--0.0069, and its
alarm rate under that shift rises from 0\% at 2{,}000 loans to 100\% at
50{,}000 loans.

\begin{table*}[t]
\centering
\caption{Boundary-calibrated alarm and detection rates (\%) over 144 looks
on 400 independent evaluation paths per setting. On separate paths at
slope 0.646 each monitor's threshold targets a 5\% alarm rate by look 144
for its observed book set. The null columns use constant generating
slopes on the base book sequence. P columns give the share whose first alarm is at or after onset,
with median delay in months among detections in brackets. N columns
give the overall alarm rate followed by that same post-onset summary.
Dashes indicate zero detections.}
\label{tab:bakeoff}
\footnotesize
\setlength{\tabcolsep}{3pt}
\begin{tabular}{@{}lccccccc@{}}
\toprule
& \multicolumn{2}{c}{Null slope} & \multicolumn{3}{c}{P1 terminal slope} & \multicolumn{2}{c}{P2 share permuted} \\
\cmidrule(lr){2-3}\cmidrule(lr){4-6}\cmidrule(l){7-8}
Monitor & 0.646 & 0.70 & 0.85 & 0.70 & 0.55 & 0.10 & 0.20 \\
\midrule
Restarted slope sum & 2.50 & 0.00 & 0.00 [---] & 0.00 [---] & 99.00 [45] & 0.00 [---] & 0.00 [---] \\
Restarted level sum & 8.00 & 6.75 & 5.75 [34] & 8.25 [21] & 7.00 [28] & 2.25 [22] & 2.75 [23] \\
Sharp-null AUC LR & 3.00 & 0.00 & 0.00 [---] & 0.00 [---] & 0.00 [---] & 0.00 [---] & 0.00 [---] \\
AUC control chart & 7.00 & 6.25 & 4.75 [80] & 4.00 [78] & 4.25 [79] & 6.50 [80] & 8.00 [40] \\
Score CUSUM & 4.50 & 10.75 & 98.50 [19] & 100.00 [17] & 100.00 [16] & 91.50 [20] & 99.75 [18] \\
Calibration test & 5.00 & 4.75 & 0.75 [19] & 4.50 [17.5] & 14.50 [17.5] & 0.75 [17] & 0.50 [15] \\
Proper-loss e-detector & 8.00 & 2.25 & 0.00 [---] & 0.75 [85] & 2.25 [84] & 0.00 [---] & 0.00 [---] \\
mROC gap & 3.25 & 0.50 & 0.00 [---] & 0.00 [---] & 31.00 [56] & 0.00 [---] & 7.00 [59] \\
\midrule
\end{tabular}\par\smallskip
\begin{tabular}{@{}lccc@{}}
Monitor & N1 +0.6 & N1 +1.2 & N2 case mix \\
\midrule
Restarted slope sum & 0.00 / 0.00 [---] & 0.00 / 0.00 [---] & 0.00 / 0.00 [---] \\
Restarted level sum & 100.00 / 99.75 [12] & 100.00 / 98.00 [7] & 3.00 / 2.25 [17] \\
Sharp-null AUC LR & 0.00 / 0.00 [---] & 0.00 / 0.00 [---] & 0.00 / 0.00 [---] \\
AUC control chart & 6.75 / 2.75 [80] & 5.25 / 0.75 [79] & 16.50 / 16.50 [0] \\
Score CUSUM & 80.75 / 80.75 [21] & 81.00 / 81.00 [21] & 74.25 / 74.25 [26] \\
Calibration test & 0.00 / 0.00 [---] & 0.00 / 0.00 [---] & 1.50 / 1.50 [15] \\
Proper-loss e-detector & 100.00 / 100.00 [21] & 100.00 / 100.00 [14] & 0.75 / 0.75 [80] \\
mROC gap & 65.25 / 65.25 [54] & 100.00 / 100.00 [19] & 0.00 / 0.00 [---] \\
\bottomrule
\end{tabular}
\end{table*}

\subsection{Quadratic calibration}\label{sec:curvature}

A quadratic conditional logit allows the slope to vary across scores.
For $x=\theta-c$, with $c=-4.35488$ the maintained commissioning
median log-odds, $\beta$ in $\beta x+\gamma x^2$ is the slope at $c$.
The monthly likelihood-ratio tests reject the linear form at nominal
5\% in 16 of 144 maintained development months and 19 of 116 holdout
months. A restricted cubic spline gives 22 and 21 rejections. Median
$|\gamma|$ ranges from 0.025 to 0.085 across the eight paths
(Table~\ref{tab:conditional-adequacy}).

The holdout decile calibration slopes on the stated scores are 0.951
for maintained and 0.773 for frozen2015. They are 1.012 and 1.006
after in-sample linear fitting, and 1.017 and 0.995 after quadratic fitting.

A case-mix shift under a quadratic link can move the linear slope
projection. At the closest tested nonzero curvature, $\gamma=0.10$ in
the panel's log-odds units, a half-standard-deviation score drift gives
95.6\% every-look coverage for the prefix projection and 79.2\% for
the fixed terminal projection. The restarted slope sum with its
simulated 120-look threshold (log 6.58) alarms on 2.2\% of paths at the
boundary, versus 4.2\% under a linear link. Table~\ref{tab:curvature-simulation}
includes the stronger curvature settings.

The restricted cubic spline uses knots at the monthly 10th, 35th, 65th
and 90th score percentiles. In the quadratic simulations, centering the
square by its book mean changes the intercept, and translating the
score origin to $c$ changes the linear coefficient by $2\gamma c$
while preserving $\gamma$. The prefix projection maximizes expected
cumulative conditional log likelihood through that look. At $\gamma=0.30$
with the same drift, the boundary alarm rate is 7.8\% and every-look
prefix coverage is 88.2\% (Table~\ref{tab:curvature-simulation}).

To test $\beta_t\ge\beta_0$ under the quadratic conditional model, let
$p_{b,g}(S_t\mid M_t)$ denote its event-set probability. We use
\[
  \frac{q_t(S_t\mid M_t)}
       {\sup_{b\ge\beta_0,\,g\in\mathbb R}p_{b,g}(S_t\mid M_t)},
\]
where $q_t$ mixes the lower-slope alternatives with curvature set to
the median of the previous 12 monthly fits. The supremum bounds the
true null probability for every event set, giving conditional
expectation at most one and the same crossing bound for the
fixed-start product. For the empirical comparison we approximate the
profile numerically. The unconstrained fit uses a gradient tolerance
of $2\times10^{-5}$ and rejects a returned gradient norm above $10^{-3}$.
When the fit lies below the boundary we maximize over curvature at
the boundary with a scalar tolerance of $10^{-9}$.
At threshold 20, the restarted slope sum with curvature profiled out
crosses on the same three development paths as with linear calibration,
and two cross later. With lagged curvature fixed, the restarted slope
sum also crosses on frozen2015 in April 2024
(Table~\ref{tab:conditional-adequacy}).

\subsection{Dependence and score scale}

A common multiplier on all loans' odds enters the intercept. At the
simulated log thresholds (2.07 for the restarted slope sum and $-0.21$
for the restarted level sum), common-shock alarm rates are 4.8--7.4\%
for the restarted slope sum and 63.8--100\% for level. With group-level
frailty up to a scale of 0.40, which can change the arrangement law, the
restarted slope sum's rates range from 3.6\% to 6.4\%.

\subsection{Vintage-level slope estimates}\label{sec:attribution}

\begin{figure*}[t]
\centering
\includegraphics[width=0.94\textwidth]{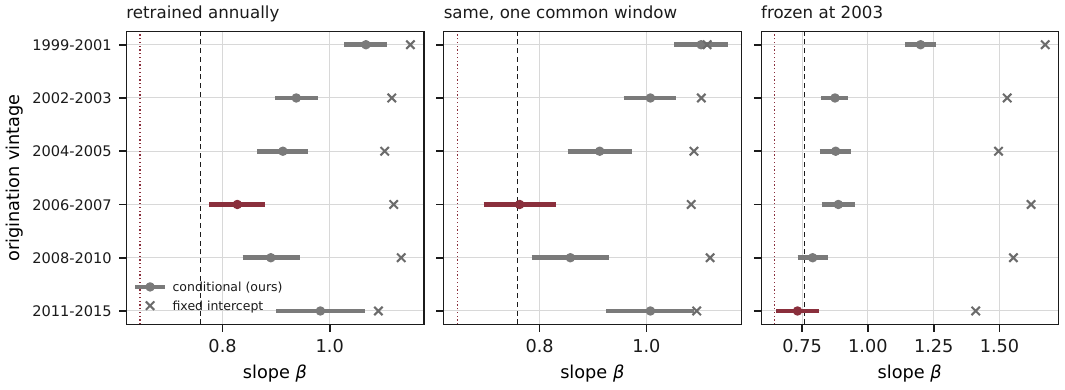}
\caption{Exploratory vintage-specific slope fits and mixture-likelihood
bands. The bands use $\alpha/6$ per vintage. Crosses mark fixed-intercept
fits. The middle panel uses the common 2011--2015 calendar window.}
\label{fig:segments}
\end{figure*}

Conditioning on the event count within each vintage gives a conditional
slope fit for that segment. In the exploratory analysis of the maintained
model, the 2006--07 vintage has a fitted slope of 0.828 versus 1.068 for
1999--2001 (Figure~\ref{fig:segments}). Restricting the analysis to the
common 2011--2015 calendar window gives 0.762 and 1.103. Across the first
five vintages in that window, monthly prepayment rates range from 1.73\%
to 1.98\%. Across the six maintained-model vintages in the full
development window, the mean fixed-intercept slope fit has a Pearson
correlation of $+0.84$ with the mean monthly prepayment rate.

For frozen2003 the fit gives 0.732 on the newest cohorts and 1.200 on the
oldest. A shared affine-logit recalibration multiplies all segment slopes
by the same factor. The newest and oldest cohorts would therefore
need separate adjustments to bring both fitted slopes to one.

\subsection{Monitoring design by book size}\label{sec:oc}

\begin{figure*}[t]
\centering
\includegraphics[width=0.86\textwidth]{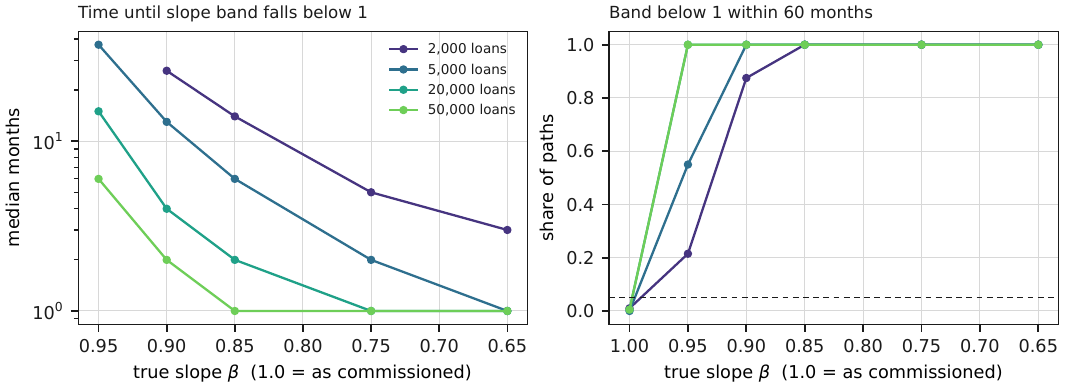}
\caption{Simulated time until the upper end of the slope band falls
below 1 (left) and share of paths meeting that criterion within 60
months (right). Dashed line marks the 5\% reference rate.}
\label{fig:oc}
\end{figure*}

We simulate 200 paths of 60 months per book size and slope
(Figure~\ref{fig:oc}), using a fixed book within each size and an intercept
chosen to keep the mean event rate fixed across slopes. At a slope of
0.90, the median time among detections to place the entire band below 1
is 26, 13, 4 and 2 months for books of 2{,}000, 5{,}000, 20{,}000 and
50{,}000 loans. At a slope of 0.85, placing the entire band below 0.90
takes a median of 40, 15 and 6 months on the three larger books. On the
2{,}000-loan book 16\% of paths meet that criterion within 60 months.
At a slope of 1 the corresponding share below 1 is 0--1\% across book
sizes, and median final band widths range from 0.158 to 0.032.

\subsection{Book-size and scale sensitivity}

\textbf{Monitored book size.}
The AUC excess share is 20\%, 15\% and 29\% on 48 development months
using 20{,}000, 60{,}000 and 120{,}000 loans, respectively. On a
12-month subset using the full risk sets of about 200{,}000 loans, the
share is 11\%.

\textbf{Rank-preserving score transforms.}
Doubling or halving the logits produces slope fits of 0.50 and 2.00.
The nonlinear monotone transform gives a fit of 0.64. Realized AUC is
unchanged across the transforms. Doubling the logits at month 19
triggers the intersection alarm that month. This sensitivity matters at
refits, while a frozen scoring function keeps its score scale.

\textbf{Attenuation.}
The monitoring rule uses the absolute tolerance $\delta=0.10$ slope
units. In the exploratory score-noise sweep, the estimated absolute gap
between commissioning and monitoring slopes shrinks from 0.149 to
0.072 at matched event intensity for a true gap of 0.15. Across noise
levels and event-intensity settings the estimated ratio of monitoring to
commissioning slopes ranges from 0.845 to 0.860 for a true ratio of 0.85.

\section{Actual-book null calibration}\label{app:panel-calibration}

For each run we preserve the observed monthly scores and event counts.
Independent draws from the conditional Bernoulli law at a constant
slope use the same suffix recursion as the Table~\ref{tab:battery}
benchmark. At the boundary, we calibrate the restarted slope sum by ordering
the maxima of 1{,}024 complete redraws. The threshold is the
$\lceil0.95(1{,}024+1)\rceil$-th ordered maximum, with a crossing
defined by strict exceedance. Each null estimate uses 1{,}024
independent validation paths. These experiments condition on the
full realized book and count sequence and describe alarm probabilities
over that sequence's horizon.

The restarted slope sum crosses the calibrated cutoff in March 2005
for frozen2003 and March 2006 for the GLM. On frozen2006 the restarted
slope sum crosses at threshold 20 in August 2011. Its peak log sum of 4.83
stays below the calibrated log cutoff of 5.46. In boundary redraws of
the 144-month runs, the restarted slope sum at 20 crosses on
49.2--57.8\% of paths. At the calibrated cutoffs, alarm fractions on
independent redraws range from 3.8\% to 5.4\% across all runs
(Table~\ref{tab:panel-calibration}). It stays below the calibrated
cutoff on both observed holdout paths (Table~\ref{tab:panel-alarms}).

To assess the restart-intersection rule with continuous sets, we evaluate the conditional
likelihood at 449 equally spaced slope values and at the generating
slope. The intersection is nonempty if any evaluated point is in
all restarted sets. For other looks the minimum of the restarted
log-likelihood margins is concave. Its maximum can exceed the best
grid value only in one of the two adjacent cells. In each cell we
bound the individual margins below by their chords and above by
extrapolating neighboring secants. Overlap of the lower bounds
certifies a surviving slope, and disjoint upper bounds certify an
empty intersection. A range in Table~\ref{tab:panel-calibration}
counts ambiguous paths once as alarm-free and once as alarmed.
These ranges bracket the simulated fractions. They are distinct from
Monte Carlo uncertainty.

\begin{table*}[t]
\centering
\caption{First observed crossings from fresh starts. Both slope
statistics test the common boundary 0.646. The fixed-start slope product,
restarted slope sum, and restarted level sum use threshold
20. The calibrated column uses the log thresholds in
Table~\ref{tab:panel-calibration}. Dashes indicate an alarm-free run.}
\label{tab:panel-alarms}
\footnotesize
\setlength{\tabcolsep}{5pt}
\begin{tabular}{@{}lccccc@{}}
\toprule
& & \multicolumn{3}{c}{Slope} & Level\\
\cmidrule(lr){3-5}\cmidrule(l){6-6}
Path & Start & Product & Sum at 20 & Calibrated sum & Sum at 20\\
\midrule
\multicolumn{6}{@{}l}{Development} \\
maintained & 2004-01 & --- & --- & --- & 2004-01 \\
frozen2003 & 2004-01 & --- & 2005-03 & 2005-03 & 2004-01 \\
frozen2006 & 2007-01 & --- & 2011-08 & --- & 2007-10 \\
frozen2009 & 2010-01 & --- & --- & --- & 2010-01 \\
frozen2012 & 2013-01 & --- & --- & --- & 2013-11 \\
GLM & 2004-01 & 2007-05 & 2005-12 & 2006-03 & 2004-01 \\
\midrule
\multicolumn{6}{@{}l}{Holdout} \\
maintained & 2016-01 & --- & --- & --- & 2018-01 \\
frozen2015 & 2016-01 & --- & --- & --- & 2018-01 \\
\bottomrule
\end{tabular}
\end{table*}

\begin{table*}[t]
\centering
\caption{Null probability estimates (\%) on the actual books and
counts. Validation of the restarted slope sum uses the boundary slope. Each
intersection column uses the corresponding constant slope. The
mean time to false alarm of the continuous-set implementation is at least 20
months under the constant-slope model. Ranges reflect numerical
bracketing. Each cell uses 1{,}024 independent validation paths.}
\label{tab:panel-calibration}
\footnotesize
\setlength{\tabcolsep}{4pt}
\begin{tabular}{@{}lrrrrrrr@{}}
\toprule
& & \multicolumn{3}{c}{Restarted slope sum} & \multicolumn{3}{c}{Intersection null slope}\\
\cmidrule(lr){3-5}\cmidrule(l){6-8}
Path & Months & Log cutoff & At 20 & At cutoff & 0.646 & 0.759 & 1.00\\
\midrule
\multicolumn{8}{@{}l}{Development} \\
maintained & 144 & 5.74 & 52.5 & 3.9 & 5.66 & 6.15--6.25 & 7.42--7.52 \\
frozen2003 & 144 & 5.39 & 49.2 & 5.4 & 5.47 & 6.45--6.54 & 6.35 \\
frozen2006 & 108 & 5.46 & 43.7 & 4.8 & 3.03 & 3.12 & 2.54 \\
frozen2009 & 72 & 4.68 & 29.1 & 5.2 & 1.46 & 1.95 & 1.17 \\
frozen2012 & 36 & 4.00 & 15.4 & 5.4 & 0.59 & 0.29 & 0.49 \\
GLM & 144 & 5.61 & 57.8 & 5.0 & 6.05--6.15 & 6.74 & 8.40--8.50 \\
\midrule
\multicolumn{8}{@{}l}{Holdout} \\
maintained & 116 & 5.76 & 49.3 & 3.9 & 4.20 & 4.10 & 3.32 \\
frozen2015 & 116 & 5.21 & 43.7 & 3.8 & 0.98 & 1.56 & 1.86--1.95 \\
\bottomrule
\end{tabular}
\end{table*}

\end{document}